%% file: MICCAI2026-main_conference_paper.tex
\documentclass[runningheads]{llncs}
\usepackage[T1]{fontenc}
\usepackage{graphicx,verbatim}
\usepackage{amsmath,amssymb}
\usepackage{booktabs}
\usepackage{tabularx}

\usepackage{booktabs}
\usepackage{multirow}
\usepackage{siunitx}
\usepackage{xurl}
\usepackage{hyperref} 
\usepackage{marvosym}
\usepackage{amsmath}

\begin{document}
\title{Benchmarking Pathology Foundation Models for Spatial Domain Understanding}


\author{
Bokai Zhao$^\text{1,2,3,4,*}$, 
Yiyang Zhang$^\text{2,3,*}$, 
Yuanchi	Zhu$^\text{2,3,5,*}$, 
Hanqing Chao$^\text{4}$, 
Long Bai$^\text{4}$, 
Tai Ma$^\text{4}$, 
Minfeng Xu$^\text{4}$, 
Ming Song$^\text{1,2,3}$, 
and Tianzi Jiang$^\text{1,2,3,5}$\textsuperscript{\Letter} 
}
\authorrunning{Zhao et al.}
\institute{$^\text{1}$School of Artificial Intelligence, University of Chinese Academy of Sciences. \\
$^\text{2}$Brainnetome Center, Institute of Automation, Chinese Academy of Sciences. \\
$^\text{3}$Beijing Key Laboratory of Brainnetome and Brain-Computer Interface, Institute of Automation, Chinese Academy of Sciences.\\
correspondence author {\Letter}: \email{jiangtz@nlpr.ia.ac.cn} \\
$^\text{4}$DAMO Academy, Alibaba Group.   $^\text{5}$ShanghaiTech University.\\
}




\maketitle              

\begin{abstract}

Pathology foundation models (PFMs) have emerged as a core approach for learning transferable representations from whole slide images (WSIs), and they are typically benchmarked through downstream clinical endpoints. While such task level evaluations are indispensable, they offer limited insight into what the representations themselves encode, particularly whether PFM embeddings can distinguish meaningful tissue regions and capture their spatial relationships. We present SpaPath-Bench, a representation level benchmark designed to diagnose spatial representation capability in PFMs. SpaPath-Bench formulates spatial domain identification (SDI) on paired whole slide image and spatial transcriptomics (ST) data as a diagnostic task. It curates 42 public paired WSI and ST slides, enables large scale evaluation across 19 encoders and seven SDI methods, and measures partition quality using three complementary criteria: unsupervised spatial coherence, transcriptomics referenced agreement, and expert referenced agreement. Across 83K runs, SpaPath-Bench reveals that different pretraining paradigms capture distinct aspects of tissue spatial architecture, and it provides practical guidance for building the next generation of spatially aware computational pathology models. Code and data pipelines are publicly available at \url{https://bokai-zhao.github.io/SpaPath-benchboard/}.


\end{abstract}

\section{Introduction}

Computational pathology develops algorithms that interpret whole-slide images (WSIs) to support diagnosis, prognosis, and biomarker discovery at scale. Beyond recognizing cellular morphology, many diagnostic and prognostic cues arise from how multiple tissue components form coherent regions and interact across a section—for example, epithelial versus stromal areas, necrotic regions, or immune-rich neighborhoods. The ability to separate meaningful tissue regions and to capture their spatial neighborhood structure is therefore a foundational capability that underpins a wide range of clinical and research applications \cite{song2023artificial}.

Pathology foundation models (PFMs) have recently become the dominant approach for learning transferable WSI representations \cite{xiong2025survey}. By self-supervised pre-training on large collections of unlabeled histology, models such as UNI \cite{UNIv1}, Virchow \cite{Virchow}, and GigaPath \cite{GigaPath} provide strong generic image embeddings. Vision–language approaches, including CONCH \cite{Conch} and MUSK \cite{MUSK}, further aim to enrich these embeddings by aligning histological patterns with medical text. Accordingly, current benchmarking efforts largely focus on downstream endpoints, e.g., mutation prediction, survival modeling, and cancer subtyping \cite{pathobench_mahmood,nc2025clinical_bench,NBME2025benchmarking}. These task-level evaluations are indispensable as they reflect practical use cases. However, they do not directly characterize what the representation itself encodes\cite{pham2023robust}: before adding task-specific heads and training aggregators, to what extent do the embeddings themselves separate meaningful tissue regions and respect their spatial continuity within a section?

We propose to answer this question using paired WSI–spatial transcriptomics (ST) datasets, where gene expression is profiled at spatially indexed spots whose coordinates are registered to the matched pathology image from the same section\cite{spatialomics2023dawn}. A common first step in ST analysis is spatial domain identification (SDI), which partitions a section into contiguous tissue regions that share similar expression profiles\cite{yuan2024benchmarking,wang2025benchmarking,liu2025high}. SDI is widely used because the inferred regions often correspond to coherent anatomical or functional  tissue regions, and many SDI methods further incorporate the matched pathology image as a morphological \cite{rao2021exploring}. This makes SDI a natural diagnostic for pathology representations: if PFM image embeddings capture meaningful tissue semantics, then spot-level embeddings extracted from the matched WSI should support SDI-style partitioning into spatially coherent regions that are biologically or anatomically meaningful regions.

Concretely, for each paired sample we extract spot-level PFM embeddings at the ST spot coordinates and treat embedding dimensions as “pseudo-expression” variables, forming a spot-by-feature matrix analogous to the gene-expression matrix used by SDI. We then apply established SDI pipelines on this pseudo-expression matrix to obtain a spatial partition. We evaluate the resulting partition using three parallel criteria: (i) unsupervised spatial coherence metrics that quantify whether inferred regions are spatially smooth and well-structured; (ii) transcriptomics-referenced agreement, comparing embedding-derived partitions against SDI partitions obtained from gene expression as a molecularly grounded target; and (iii) expert-referenced agreement, comparing against manual anatomical annotations when available. Together, these evaluations test whether PFM embeddings support both region separability and spatial neighborhood consistency, and whether the resulting partitions align with transcriptomics-derived structure or expert tissue labels.

Based on this view, we introduce \emph{SpaPath-Bench}, a systematic benchmark for evaluating spatial organization encoded in pathology foundation models. Our contributions are threefold: 

- We formulate SDI on PFM image embeddings as a representation-level benchmark that complements conventional downstream task-level evaluation. 

- We provide a three-way evaluation protocol: unsupervised spatial coherence, transcriptomics-referenced agreement, and expert-referenced agreement, capturing distinct yet meaningful notions of region validity. 

- We conduct a large-scale factorial study on 42 public paired WSI–ST datasets across 19 encoders, 7 SDI methods, and multiple clustering algorithms, yielding 83k total runs with statistical reporting.

\begin{figure}[t]
\includegraphics[width=\textwidth]{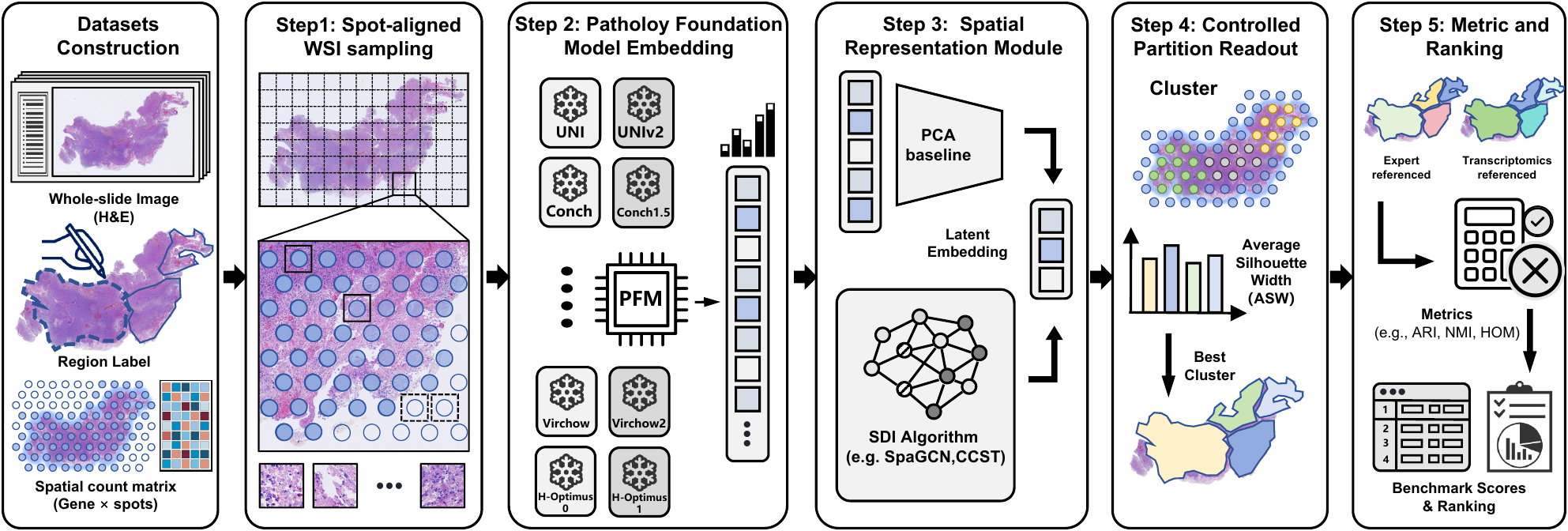}
\caption{Overview of the benchmark pipeline for spatial domain understanding.  } \label{fig_1}
\end{figure}

\section{Benchmark Protocol}

\subsection{Benchmark Pipeline}
\label{sec:pipeline}

Given a spatial transcriptomics (ST) slide with a paired H\&E image, we denote the measured spots as $\mathcal{S}=\{s_i\}_{i=1}^{N}$, where each spot $s_i$ is associated with a 2D physical coordinate $\mathbf{p}_i\in\mathbb{R}^2$ in the tissue plane. Our benchmark evaluates whether pathology foundation model (PFM) representations preserve information sufficient for recovering \emph{spatial domain structures} by standardizing the pipeline from spot-level feature extraction to domain partitioning and metric computation.

\textbf{Step 1: Spot-aligned WSI sampling.}
For each spot $s_i \in \mathcal{S}$, we extract an aligned histology patch $x_i$ from the paired H\&E image centered at $\mathbf{p}_i$ (using a fixed magnification and patch size). We apply only model-input standardization, including cropping and intensity normalization, to obtain the spot-level image inputs $\mathcal{X}=\{x_i\}_{i=1}^{N}$.

\textbf{Step 2: Foundation model embedding.}
Given a pre-trained encoder $f_{\theta}$ (a PFM or a baseline vision model), we compute a $D$-dimensional embedding for each patch as $\mathbf{z}_i=f_{\theta}(x_i)\in\mathbb{R}^{D}$. Stacking these embeddings forms the feature matrix $\mathbf{Z}=[\mathbf{z}_1,\ldots,\mathbf{z}_N]^{\top}\in\mathbb{R}^{N\times D}$. We conceptualize $\mathbf{Z}$ as a \emph{pseudo-expression} matrix, where each latent dimension acts as a \emph{virtual gene} measured at that spot, aligning histology-derived representations with standard ST analysis workflows.

\textbf{Step 3: Spatial representation module (PCA vs.\ SDI-based).}
To incorporate spatial context, we transform $\mathbf{Z}$ into a spatially informed representation $\mathbf{H}\in\mathbb{R}^{N\times d}$ via one of two routes:
\textbf{(i) Non-spatial baseline (PCA):} $\mathbf{H}=\mathrm{PCA}_d(\mathbf{Z})$ with $d\in\{32,64,128,256\}$.
\textbf{(ii) SDI-based spatial representation:} We construct a spatial neighborhood graph $G=(V,E)$ over $\mathcal{S}$ using $k$-nearest neighbors ($k=6$) based on the coordinate matrix $\mathbf{P}=[\mathbf{p}_1,\ldots,\mathbf{p}_N]^{\top}\in\mathbb{R}^{N\times 2}$. An SDI method $g_{\phi}$ then learns the spatial representations as $\mathbf{H}=g_{\phi}(\mathbf{Z},\mathbf{P},G)$, acting as a standardized \emph{spatial representation learner} operating on the pseudo-expression space to promote neighborhood coherence.

\textbf{Step 4: Controlled partition readout (clustering).}
Given $\mathbf{H}$, we generate a spatial domain partition $\hat{\mathbf{y}}\in\{1,\ldots,K\}^{N}$ using a controlled clustering function $c \in \{\text{K-means},\,\text{Louvain},\,\text{Leiden}\}$, i.e., $\hat{\mathbf{y}}=c(\mathbf{H})$. For expert-annotated slides, $K$ matches the number of reference domains; for unlabeled slides, $K$ is determined via reference-free criteria (Sec.~\ref{sec:metrics}). Decoupling \emph{representation learning} (PCA/SDI) from the \emph{partition readout} (clustering) makes performance differences more directly attributable to the learned embeddings under consistent readout choices.

\textbf{Step 5: Metric computation and ranking.}
We quantify spatial domain fidelity by comparing $\hat{\mathbf{y}}$ against available references (transcriptomics-referenced agreement and expert-referenced agreement), alongside reference-free spatial coherence metrics (Sec.~\ref{sec:metrics}). Results are aggregated across slides and three independent random seeds to report robust summary statistics and rankings.

\subsection{Evaluation Metrics Protocol}
\label{sec:metrics}
We quantify spatial domain understanding via three complementary regimes: (i) structural alignment with expert knowledge, (ii) biological concordance with molecular profiles, and (iii) intrinsic spatial coherence.

\noindent\textbf{Reference-free Metrics.} To assess intrinsic spatial quality without any external labels, we compute: (i) CHAOS, to measure spatial dispersion; (ii) Patch Adjacency Score (PAS), to evaluate local neighborhood consistency; and (iii) Average Silhouette Width (ASW), to quantify cluster separation in the latent representation space.

\noindent\textbf{Reference-based Metrics.} Given a predicted partition $\hat{\mathbf{y}}$ and a reference $\mathbf{y}^{\ast}$, we compute Adjusted Rand Index (ARI), Normalized Mutual Information (NMI), homogeneity (HOM), and completeness (COM). 

 \textit{(1) Expert-referenced (labeled slides):} For slides with manual annotations, we set the cluster number to the ground truth, $K=|\mathcal{Y}^{\text{expert}}|$, and compute agreement against expert labels. 
 
 \textit{(2) Transcriptomics-referenced (unlabeled slides):} For slides lacking manual labels, we determine $\hat{\mathbf{y}}$ by maximizing the ASW over $K\in\{4,\dots,12\}$. A proxy reference $\mathbf{y}^{\text{HVG}}$ is independently constructed by applying the same SDI and clustering pipeline to the top 2000 highly variable genes (HVGs), with its $K$ similarly selected via maximum ASW. Agreement between $\hat{\mathbf{y}}$ and $\mathbf{y}^{\text{HVG}}$ quantifies the alignment between morphology-derived patterns and molecular-defined functional domains.

\noindent\textbf{Rank-score and Statistical Robustness.} To ensure robust comparison across 3 random seeds and 42 slides, we perform one-sided Wilcoxon signed-rank tests under matched configurations (same SDI method, clustering readout, and seed). A model $m_i$ earns a rank-score $S(m_i,c)=\sum_{j\neq i}\mathbb{I}(p_{i,j,c}<0.05)$, representing the number of competitors it significantly outperforms. Overall benchmark rankings are obtained by aggregating $S(m_i,c)$ across all configurations and slides.

\section{Experiments}
\subsection{Benchmark Datasets Construction}
\label{sec:data}

We curate a benchmark of 42 publicly available spatial transcriptomics (ST) slides with paired H\&E images, comprising one expert-annotated cohort and one unlabeled cohort.

- \textit{\textbf{Expert-annotated cohort.}}
We include 12 human dorsolateral prefrontal cortex (DLPFC) slides \cite{huuki2024data} with spot-level expert annotations into 7 anatomical regions (Layers 1-6 and White Matter). This cohort enables reference-based evaluation of domain recovery against human-interpretable anatomical boundaries.

- \textit{\textbf{Unlabeled HE--ST cohort.}}
We collect 30 additional slides without expert annotations, spanning multiple tissues and both human and mouse samples. For these slides, we quantify morphology--molecule consistency using the omics-guided proxy evaluation protocol (Sec.~\ref{sec:metrics}), comparing morphology-derived partitions with transcriptomics-derived HVG reference partitions.

\input{table-model}

\subsection{Experimental Setup}
\label{sec:exp_setup}

\noindent\textbf{Foundation Models (Encoders).} 
We systematically evaluate 19 foundation models, categorizing them into four distinct pre-training paradigms: (i) \emph{Pure Vision} PFMs (e.g., UNI/UNI2\cite{UNIv1}, Virchow/Virchow2\cite{Virchow}, H-Optimus-0/1, GigaPath\cite{GigaPath}, Phikon/Phikon2\cite{filiot2024phikonv2largepublicfeature}, Hibou-B/L\cite{nechaev2024hibou}, CtransPath\cite{wang2024pathology}); (ii) \emph{Vision-Language} PFMs (CONCH/ CONCH-V1.5\cite{Conch}, PLIP\cite{huang2023visual}, MUSK\cite{MUSK}), which align morphological features with medical text; (iii) \emph{Vision-Omics} models (OmicsCLIP\cite{OmicsCLIP}), bridging histology and molecular profiles; and (iv) \emph{Natural Image} baselines (DINOv3-ViT-B/L\cite{simeoni2025dinov3}). Detailed specifications of each model, including architecture, parameter count, and pre-training corpus, are summarized in Table~\ref{tab:models}.

\noindent\textbf{SDI Methods and Clustering.} 
The high-dimensional representations extracted by the frozen PFMs are processed through a unified spatial module. We compare conventional dimensionality reduction (PCA) against 7 mainstream Spatial Domain Identification (SDI) algorithms, systematically categorized by their core learning paradigms: (i) \emph{Graph Clustering} (SpaGCN\cite{hu2021spagcn}, CCST\cite{CCST}); (ii) \emph{Graph Autoencoders} (SEDR\cite{xu2024unsupervised}, STAGATE\cite{dong2022deciphering}); (iii) \emph{Contrastive Learning} (conST\cite{zong2022const}, GraphST\cite{long2023spatially}); and (iv) \emph{Manifold Learning} (SpaceFlow\cite{ren2022identifying}). Unlike PCA, these graph-based methods explicitly model tissue topology by constructing spatial adjacency graphs from spot coordinates. The resulting spatially-aware embeddings are partitioned into functional domains using three clustering strategies: K-means, Louvain\cite{wolf2018scanpy}, and Leiden\cite{traag2019louvain}. To ensure a fair and reproducible benchmark, we strictly adhere to the default hyperparameters (e.g., learning rates, latent dimensions, training epochs) recommended by the respective original publications. All detailed configurations are comprehensively summarized in Table~\ref{tab:sdi_methods}, and all evaluations are rigorously repeated across 3 independent random seeds.
\input{table-SDI}

\begin{figure}[t]
\includegraphics[width=\textwidth]{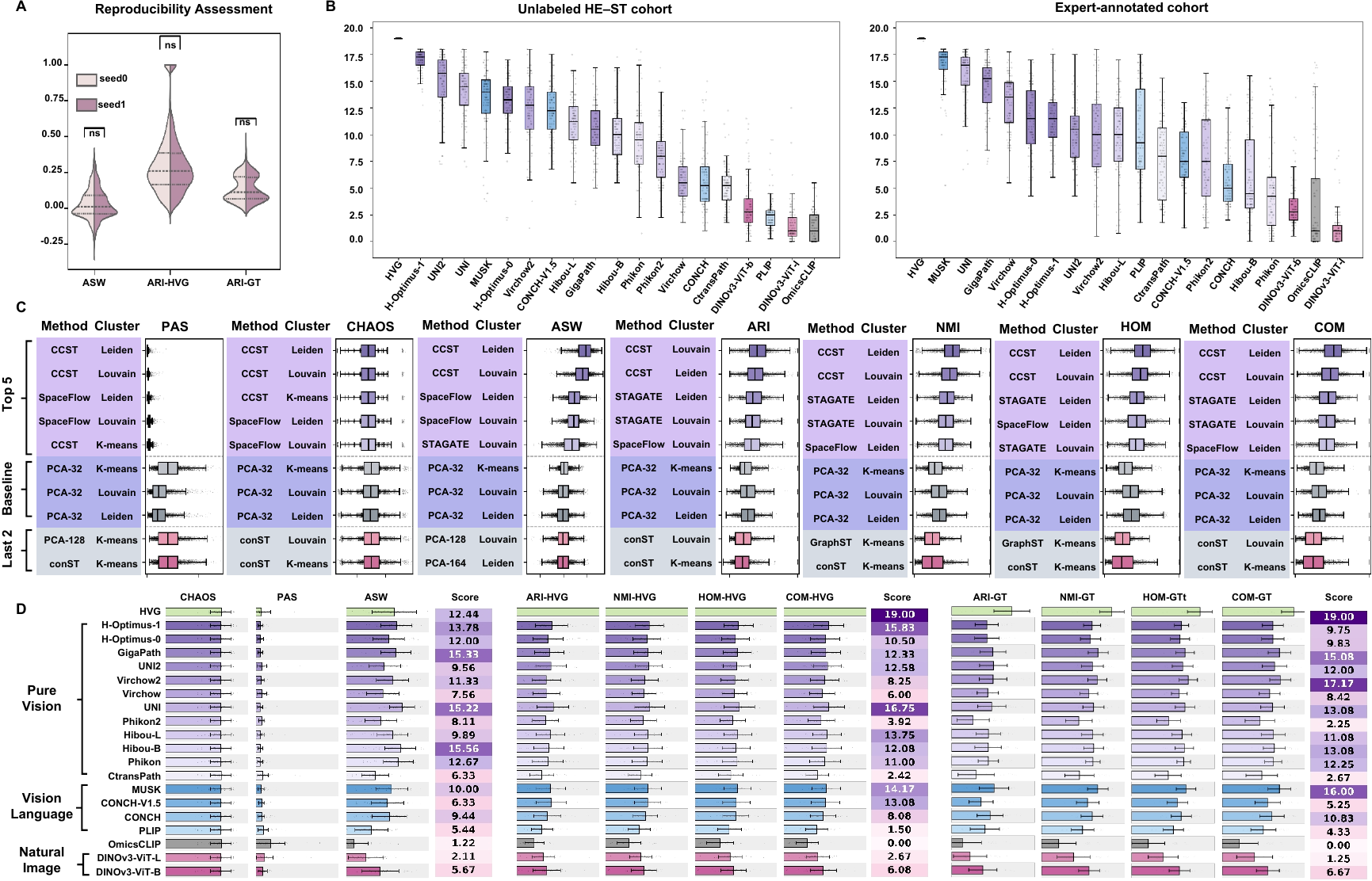}
\caption{Overall benchmark summary evaluating PFM spatial domain understanding.  } \label{fig_2}
\end{figure}

\section{Results}
\subsection{Overall Benchmark Summary}
\label{subsec:overall_summary}

\noindent\textbf{Pipeline Stability (Fig.\ref{fig_2}A).} We evaluated the stability of our pipeline by comparing metric distributions across three independent random seeds. Paired Wilcoxon signed-rank tests revealed no statistically significant differences ($p > 0.05$) across all evaluation metrics, indicating that the benchmark outcomes are not driven by stochastic initialization.

\noindent\textbf{Global Model Rankings (Fig.\ref{fig_2}B).} Aggregating evaluation metrics across all 42 ST slides and all clustering methods, we established a global ranking for the 19 evaluated models. When assessed against the transcriptomic proxy (HVG-guided reference), \noindent\textbf{H-Optimus-1} achieved the highest overall rank. Conversely, when evaluated against expert manual annotations (the DLPFC subset), \textbf{MUSK} recorded the highest performance metrics.

\noindent\textbf{Efficacy of Spatial Representations (Fig.\ref{fig_2}C).}
We systematically compared diverse combinations of SDI methods and clustering algorithms.
For each metric, we visualize the top five configurations alongside three PCA baselines and the two lowest-ranked methods ; among all evaluated settings, \textit{CCST+Leiden} yields the best aggregate performance.

\noindent\textbf{Model Rankings under Optimal Setting (Fig.\ref{fig_2}D).} As a specific case study utilizing the optimal \textit{CCST+Leiden} configuration, we summarized the evaluation metrics across all 42 ST slides to present the detailed performance ranking of the 19 PFMs under this high-performing setting.

\begin{figure}[t]
\includegraphics[width=\textwidth]{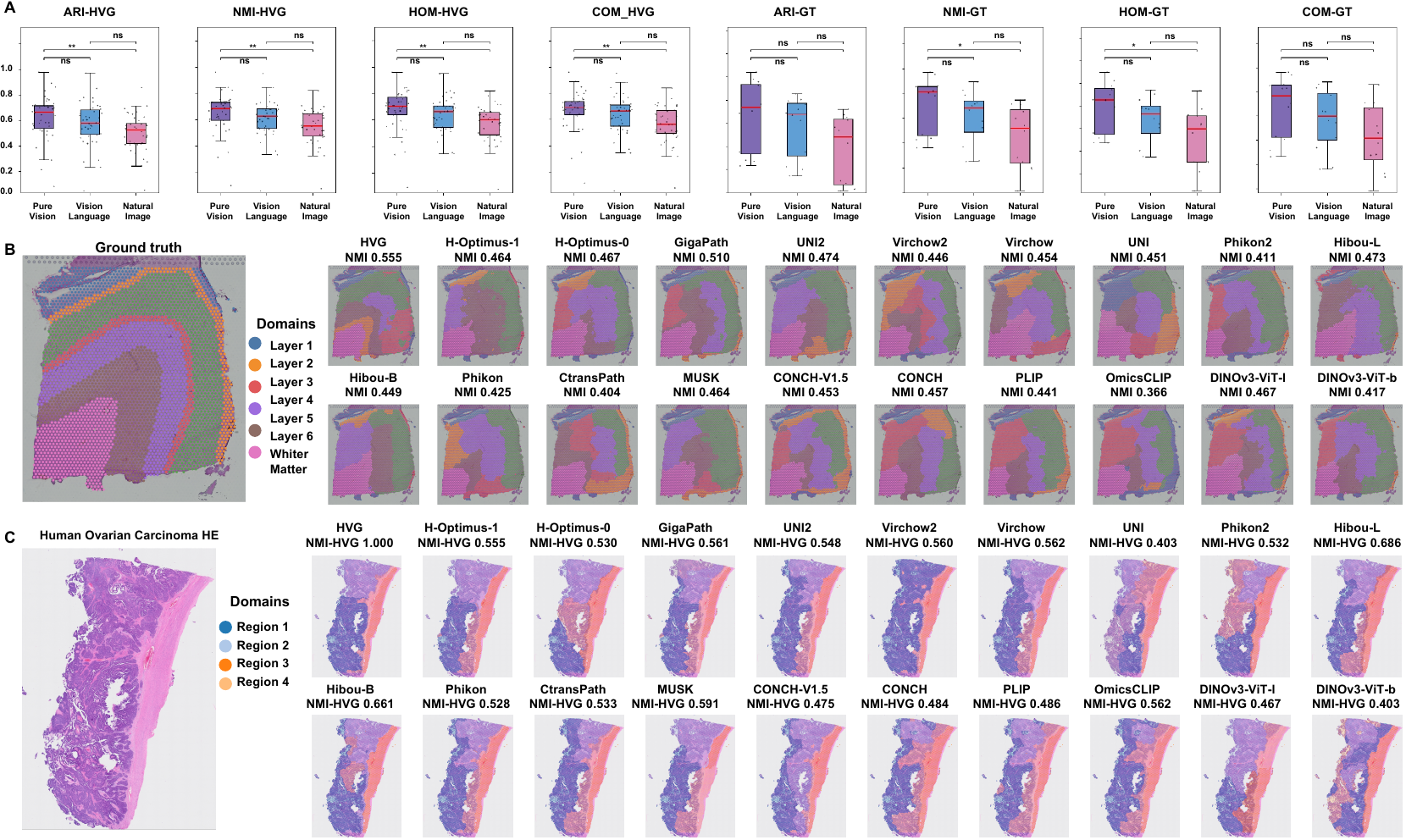}
\caption{Pre-training paradigm impacts (A) and qualitative spatial domain visualizations on DLPFC (B) and ovarian carcinoma (C) slides.} \label{fig_3}
\end{figure}

\subsection{Key Findings}
\label{subsec:findings}
\textbf{Reference-dependent representations.} \textbf{H-Optimus-1} shows strong performance when evaluated against the transcriptomic proxy (HVG), implying self-supervised learning can capture fine-grained molecular heterogeneity. Conversely, \textbf{MUSK} achieves top results on expert annotations, suggesting vision-language pre-training may better align with human-defined macro-architectural semantics. \textbf{WSI-level spatial contrastive learning.} CCST generally outperforms baselines, indicating that integrating whole-slide contrastive learning helps aggregate regional context for holistic tissue comprehension, improving upon isolated patch features. \textbf{Pre-training paradigm impacts.} (Fig.\ref{fig_3}A) While pathology models tend to outperform natural-image baselines, vision-language models do not consistently show a distinct advantage over pure-vision counterparts. Strict patch-level text-image alignment might potentially limit the macroscopic spatial coherence beneficial for tissue-level clustering.

\subsection{Qualitative Visual Assessment}
\label{subsec:qualitative}

Qualitative visualizations on two representative slides illustrate how PFM-derived embeddings translate into spatial domains under our pipeline. 
\textbf{Expert-annotated case (DLPFC, Fig.\ref{fig_3}B):} Because cortical layers exhibit subtle and gradual morphological transitions, histology-driven partitions show only partial agreement with expert annotations. This is consistent with the fact that expert layer labels encode high-level neuroanatomical context that may not be fully determined by local H\&E appearance alone. 
\textbf{Unlabeled case (Ovarian carcinoma, Fig.\ref{fig_3}C):} In contrast, for tissues with pronounced structural heterogeneity, PFM-based domains delineate distinct microenvironments and align well with visually apparent macro-scale boundaries.

\section{Summary}
We introduced \textit{SpaPath-Bench}, a standardized benchmark to assess spatial organization encoded in PFMs using SDI on spot-aligned embeddings. With a unified pipeline and three evaluation regimes (expert-referenced, transcriptomics-referenced, and reference-free spatial coherence), we systematically compare 19 encoders on 42 paired WSI--ST slides and identify strengths and bottlenecks in region separability and neighborhood coherence. We envision \textit{SpaPath-Bench} as a testbed for developing more robust, context-aware pathology representations that better integrate morphology with spatial omics.

\begin{credits}
\subsubsection{\ackname} This work was supported by Brain Science and Brain-like Intelligence Technology - National Science and Technology Major Project 2021ZD0200200.

\subsubsection{\discintname}
The authors declare that they have no competing interests.
\end{credits}

\bibliography{reference}
\bibliographystyle{splncs04}

\end{document}

%% file: table-model.tex
\begin{table}[htbp]
\centering
\caption{Summary of evaluated foundation models and pre-training details.}
\label{tab:models}
\footnotesize
\setlength{\tabcolsep}{3.0pt}
\renewcommand{\arraystretch}{0.92}
\resizebox{\columnwidth}{!}{%
\begin{tabular}{c l c c c c c c l}
\toprule
\textbf{Modality} & \textbf{PFM} & \textbf{Arch} & \textbf{Input} & \textbf{Dim} & \textbf{Params.} & \multicolumn{2}{c}{\textbf{Data}} & \textbf{Pretrain} \\
\cmidrule(lr){7-8}
& & & (px) & $D$ & & \#Slides & \#Patches & \textbf{Methods} \\
\midrule
\multirow[c]{12}{1.35cm}{\centering Pure\\Vision}
& H-Optimus-1   & ViT-G/14   & 224 & 1536 & 1.1B & 1M    & 2.0B  & -- \\
& H-Optimus-0   & ViT-G/14   & 224 & 1536 & 1.1B & 500K  & --    & DINOv2 \\
& GigaPath      & ViT-G/14   & 224 & 1536 & 1.1B & 171K  & 1.4B  & DINOv2 \\
& UNI2          & ViT-H/14   & 224 & 1536 & 681M & 350K  & 200M  & DINOv2 \\
& Virchow2      & ViT-H/14   & 224 & 1280 & 631M & 3134K & 2.0B  & DINOv2 \\
& Virchow       & ViT-H/14   & 224 & 1280 & 631M & 1488K & 2.0B  & DINOv2 \\
& UNI           & ViT-L/16   & 224 & 1024 & 304M & 100K  & 100M  & DINOv2 \\
& Phikon2       & ViT-L/16   & 224 & 1024 & 304M & 58K   & 456M  & DINOv2 \\
& Hibou-L       & ViT-L/16   & 224 & 1024 & 304M & 1100K & 1.2B  & DINOv2 \\
& Hibou-B       & ViT-B/16   & 224 & 768  & 86M  & 1100K & 1.2B  & DINOv2 \\
& Phikon        & ViT-B/16   & 224 & 768  & 86M  & 6,093 & 43.4M & iBOT \\
& CtransPath    & Swin-T/14  & 224 & 768  & 28M  & 60,530& 15.6M & MoCov3 \\
\midrule
\multirow[c]{4}{1.35cm}{\centering Vision-\\Language}
& MUSK        & ViT-L/16 & 384 & 2048 & 304M & -- & 1M    & BEIT-3 \\
& CONCH-V1.5  & ViT-L/16 & 512 & 768  & 304M & -- & 1.26M & CoCa \\
& CONCH       & ViT-B/16 & 224 & 512  & 86M  & -- & 1.17M & CoCa \\
& PLIP        & ViT-B/32 & 224 & 512  & 86M  & -- & 208K  & CLIP \\
\midrule
\multirow[c]{1}{1.35cm}{\centering V-Omics}
& OmicsCLIP   & ViT-L/14 & 224 & 768 & 304M & 1,007 & 2.2M & CoCa \\
\midrule
\multirow[c]{2}{1.35cm}{\centering Natural\\Image}
& DINOv3-ViT-l & ViT-L/16 & 256 & 1024 & 304M & -- & -- & DINOv3 \\
& DINOv3-ViT-b & ViT-B/16 & 256 & 768  & 86M  & -- & -- & DINOv3 \\
\bottomrule
\end{tabular}%
}
\end{table}

%% file: table-SDI.tex
\begin{table*}[htbp]
\centering
\caption{Summary of evaluated SDI methods and default hyperparameters.}
\label{tab:sdi_methods}
\small
\renewcommand{\arraystretch}{0.85} 
\setlength{\tabcolsep}{3.2pt}     
\resizebox{\columnwidth}{!}{%
\begin{tabular}{lccccccc}
\toprule
\textbf{Category} & \multicolumn{2}{c}{\textbf{Graph Clustering}} & \multicolumn{2}{c}{\textbf{Graph Autoencoder}} & \multicolumn{2}{c}{\textbf{Contrastive}} & \textbf{Manifold} \\
\cmidrule(lr){2-3}\cmidrule(lr){4-5}\cmidrule(lr){6-7}\cmidrule(lr){8-8}
\textbf{Method}
& \textbf{SpaGCN} & \textbf{CCST}
& \textbf{SEDR} & \textbf{STAGATE}
& \textbf{conST} & \textbf{GraphST}
& \textbf{SpaceFlow} \\
\midrule
Dims   & 50   & 256  & 32  & 30   & 28 & 28 & 50 \\
Epochs        & 2000 & 5000 & 200 & 1000 & 200 & 500 & 1000 \\
LR & 0.05 & $10^{-6}$ & 0.01 & $10^{-3}$ & 0.01 & $5\times 10^{-4}$ & $10^{-3}$ \\
\bottomrule
\end{tabular}%
}
\end{table*}